\def\year{2022}\relax
\documentclass[letterpaper]{article} 
\usepackage{aaai22}  
\usepackage{times}  
\usepackage{helvet}  
\usepackage{courier}  
\usepackage[hyphens]{url}  
\usepackage{graphicx} 
\usepackage{natbib}  
\usepackage{caption} 
\DeclareCaptionStyle{ruled}{labelfont=normalfont,labelsep=colon,strut=off} 
\usepackage{algorithm}
\usepackage{algorithmic}

\usepackage{times}
\usepackage{epsfig}
\usepackage{graphicx}
\usepackage{amsmath}
\usepackage{amssymb}
\usepackage{subfigure}
\usepackage{paralist}
\usepackage{xcolor}
\usepackage{booktabs}  
\usepackage{multirow} 
\usepackage{verbatim}
\usepackage{arydshln}
\usepackage{mathtools}
\usepackage{enumitem}

\newtheorem{theorem}{Theorem}
\newtheorem{lemma}{Lemma}

\newtheorem{definition}{Definition}

\newsavebox\CBox
\def\textBF#1{\sbox\CBox{#1}\resizebox{\wd\CBox}{\ht\CBox}{\textbf{#1}}}

\newlength\secmargin
\newlength\paramargin
\newlength\figmargin

\usepackage{newfloat}
\usepackage{listings}
\floatstyle{ruled}
\newfloat{listing}{tb}{lst}{}
\floatname{listing}{Listing}
\title{Lifelong Generative Modelling Using Dynamic Expansion Graph Model}
\author{
	Fei Ye and Adrian G. Bors
}
\affiliations{
	Department of Computer Science, University of York, York YO10 5GH, UK\\
	fy689@york.ac.uk, adrian.bors@york.ac.uk
}

\usepackage{bibentry}

\begin{document}
	
	\maketitle
	
	\begin{abstract}
		Variational Autoencoders (VAEs) suffer from degenerated performance, when learning several successive tasks. This is caused by catastrophic forgetting. In order to address the knowledge loss, VAEs are using either Generative Replay (GR) mechanisms or Expanding Network Architectures (ENA). In this paper we study the forgetting behaviour of VAEs using a joint GR and ENA methodology, by deriving an upper bound on the negative marginal log-likelihood. This theoretical analysis provides new insights into how VAEs forget the previously learnt knowledge during lifelong learning. The analysis indicates the best performance achieved when considering model mixtures, under the ENA framework, where there are no restrictions on the number of components. However, an ENA-based approach may require an excessive number of parameters. This motivates us to propose a novel Dynamic Expansion Graph Model (DEGM). DEGM expands its architecture, according to the novelty associated with each new databases, when compared to the information already learnt by the network from previous tasks. DEGM training optimizes knowledge structuring, characterizing the joint probabilistic representations corresponding to the past and more recently learned tasks. We demonstrate that DEGM guarantees optimal performance for each task while also minimizing the required number of parameters. Supplementary materials (SM) and source code are  available\footnote{ https://github.com/dtuzi123/Expansion-Graph-Model}.
	\end{abstract}

	\section{Introduction}
	
	The Variational Autoencoder (VAE) \cite{VAE} is a popular generative deep learning model with remarkable successes in learning unsupervised tasks by inferring probabilistic data representations \cite{VAE_symmetric}, for disentangled representation learning \cite{baeVAE,JontLatentVAEs} and for image reconstruction tasks \cite{MixtureOfVAEs,DeepMixtureVAE,InfoVAEGAN_conference,JontLatentVAEs}. Training a VAE model involves maximizing the marginal log-likelihood $\log p_\theta({\bf x}) = \log \int p_\theta({\bf x} | {\bf z})p({\bf z}) \mathrm{d}{\bf z}$ which is intractable during optimization due to the integration over the latent space defined by the variables ${\bf z}$. VAEs introduce using a variational distribution $q_\omega ( {\bf z} | {\bf x} )$ to approximate the posterior and the model is trained by maximizing a lower bound, called Evidence Lower Bound (ELBO), \cite{VAE}~:
	\begin{equation}
	\begin{aligned}
	&\log p_\theta ({\bf x}) \ge  {\mathbb{E}_{{q_\omega }\left( {{\bf z}\mid{\bf x}} \right)}}\left[ {\log {p_\theta }\left( {{\bf x}\,|\,{\bf z}} \right)} \right]  \\&- KL\left[ {{q_\omega }\left( {{\bf z}\,|\,{\bf x}} \right) \mid\mid p\left( {\bf z} \right)} \right] := \mathcal{L}_{ELBO}\left({\bf x} ; \{\theta,\omega \} \right) 
	\label{singleElbo}
	\end{aligned}
	\end{equation}
	where $p_\theta ({\bf x} \,|\, {\bf z})$ and $p({\bf z})=\mathcal{N}(0,I)$ are the decoding and prior distribution, respectively, while $KL[\cdot]$ represents the Kullback–Leibler divergence. Defining a tighter ELBO to the marginal log-likelihood, achieved by using a more expressive posterior \cite{RecursiveVAE,Aux_DGM}, importance sampling \cite{IWVAE,IWVAE2} or through hierarchical variational models \cite{DoublySemiVAE,NVAE}, has been successful for improving the performance of VAEs. However, these approaches can only guarantee a tight ELBO for learning a single domain and have not yet been considered for lifelong learning (LLL), which involves learning sequentially several tasks associated with different databases. VAEs, similarly to other deep learning methods \cite{ImprovedAGEM}, suffer from catastrophic forgetting \cite{Catastrophic}, when learning new tasks, leading to degenerate performance on the previous tasks. One direct way enabling VAE for LLL is the Generative Replay (GR) process \cite{GenerativeLifelong}. 
	
	Let us consider a VAE model to be trained on a sequence of $t$ tasks. After the learning of $i$-th task is finished, the GR process allows the model to generate a pseudo dataset $\tilde{\bf X}^i$ which will be mixed with the incoming data set ${\bf X}^{new}$ to form a joint dataset for the $(i+1)$-th task learning. Usually, the distribution of $\{{\tilde{\bf X}}^i,{\bf X}^{new} \}$ does not match the real data distribution exactly and the optimal parameters $\{ \theta^*,\omega^* \}$ are estimated by maximizing ELBO, on samples ${\bf x}'$ drawn from $\{{\tilde{\bf X}}^i,{\bf X}^{new} \}$. $\mathcal{L}_{ELBO}(\cdot)$ is not a tight ELBO in Eq.~\eqref{singleElbo} by using the model's parameters $\{ \theta^*,\omega^* \}$ which actually are not optimal for the real sample log-likelihood $\log p_\theta({\bf x})$ 
	(See Proposition 6 in Appendix-I from SM$^{1}$).
	In this paper, we aim to evaluate the tightness between $\log p_\theta({\bf x})$ and ${\mathcal{L}}_{ELBO}({\bf x}';\theta^*,\omega^*)$, by developing a novel upper bound to the negative marginal log-likelihood, called Lifelong ELBO (LELBO). LELBO involves the discrepancy distance \cite{domainTheory} between the target and the evolved source distributions, as well as the accumulated errors, caused when learning each new task. This analysis provides insights into how the VAE model is losing previously learnt knowledge during LLL. We also generalize the proposed theoretical analysis to ENA models, which leads to a novel dynamic expansion graph model (DEGM) enabled with generating graph structures linking the existing components and a newly created component, benefiting on the transfer learning and the reduction of the model's size. We list our contributions as~: 
	\renewcommand\labelitemi{\large$\bullet$}
	\begin{itemize}
		\item This is the first research study to develop a novel theoretical framework for analyzing VAE's forgetting behaviour during LLL.
		\item We develop a novel generative latent variable model which guarantees the trade-off between the optimal performance for each task and the model's size during LLL.
		\item We propose a new benchmark for the probability density estimation task under the LLL setting.
	\end{itemize}
	
	\section{Related works}
	
	Recent efforts in LLL focus on regularization based methods \cite{LessForgetting,Lwf}, which typically penalize significant changes in the model's weights when learning new tasks. Other methods rely on memory systems such as using past learned data to guide the optimization \cite{AGEM,ImprovedAGEM,Functional_Regularisation_LLL}, using Generative Adversarial Nets (GANs) or VAEs \cite{Lifelong_VAE,GenerativeLifelong,LifelongTwin,Generative_replay,LifelongVAEGAN,LifelongInterpretable,LifelongTeacherStudent} aiming to reproduce previously learned data samples in order to attempt to overcome forgetting. However, most of these models focus on predictive tasks and the lifelong generative modelling remains an unexplored area. 
	
	Prior works for continuously learning VAEs are divided into two branches: Generative Replay (GR) and Expanding Network Architectures (ENA). GR was used in VAEs for the first time in \cite{Lifelong_VAE} while \cite{GenerativeLifelong} extends the GR mechanism within a Teacher-Student framework, called the Lifelong Generative Modelling (LGM). A major limitation for GR is its inability of learning a long sequence of data domains. This is due to its fixed model capacity while having to retrain the generator frequently \cite{LifelongVAEGAN}. This issue is relieved by using ENA \cite{NeuralDirchlet}, inspired by a network expansion mechanism \cite{LifelongUnsupervisedVAE}, or by employing a combination between ENA and GR mechanisms \cite{LifelongMixuteOfVAEs,LifelongInfinite}. These methods significantly relieve forgetting but would suffer from informational interference when learning a new task \cite{InterferenceLifelongLearning}.
	
	The tightness on ELBO is key to improving VAE's performance and one possible way is to use the Importance Weighted Autoencoder (IWELBO) \cite{IWVAE} in which the tightness is controlled by the number of weighted samples considered. Other approaches focus on the choice of the approximate posterior distribution, including by using normalizing flows \cite{ImpVarInf,VAE_NormFlow}, employing implicit distributions \cite{AdbVB} and using hierarchical variational inference \cite{HImportedVAE}. The IWELBO bound can be used with any of these approaches to further improve their performance \cite{ImportanceHVAE}. Additionally, online variational inference \cite{VCL} has been used in VAEs, but require to store the past samples for computing the approximate posterior, which is intractable when learning an infinite number of tasks. The tightness of ELBO under LLL was not studied in any of these works.
	
	\section{Preliminary}
	
	In this paper, we address a more general lifelong unsupervised learning problem where the task boundaries are provided only during the training. For a given sequence of tasks $\{ {\mathcal T}_1,\dots,{\mathcal T}_N\} $ we consider that each ${\mathcal T}_i$ is associated with an unlabeled training set $Q_i^S$ and an unlabeled testing set $Q_i^T$. The model only sees a sequence of training sets $\{ Q_1^S,\dots,Q_N^S\}$ while it is evaluated on $\{ Q_1^T,\ldots,Q_N^T\} $. Let us consider the input data space $ \mathcal{X} \in {\mathbb R}^d$ of dimension $d$, and $\mathcal{P}_i$ the probabilistic representation of the testing set $Q_i^T$. We desire to evaluate the quality of reconstructing data samples ${\bf x} \in \mathcal{X}$, by a model using the square loss (SL) function $\|{\bf x} - h({\bf x})\|^2$, where $h$ is a hypothesis function in a space of hypotheses $\{h \in \mathcal{H} \mid \mathcal{H}: \mathcal{X} \to \mathcal{X} \}$. For the image space, the loss is represented by $\sum\nolimits_{i = 1}^d {({\bf x}[i] - h({\bf x})[i])}^2$, where $[i]$ represents the entry for the $i$-th dimension.
	
	\begin{definition} (\textBF{Single model.})
		Let ${\cal M} = \{ f_\omega,g_\theta\} $ be a single model consisting of an encoder $f_\omega \colon \mathcal{X} \to \mathcal{Z}$
		for representing $q_\omega({\bf z}\,|\,{\bf x})$, and 
		a decoder $g_\theta: \mathcal{Z} \to \mathcal{X}$ for modelling $p_\theta({\bf x}\,|\,{\bf z})$. The latent variable ${\bf z} = f_\omega^\mu ({\bf x}) + f_\omega^\delta ({\bf x}) \odot \gamma$, $\gamma \sim \mathcal{N}(0,I)$ is reparameterized by the mean $f_\omega^\mu ({\bf x})$ and variance $f_\omega^\delta ({\bf x})$, implemented by a network $f_\omega({\bf x})$. $\{ \omega^t,\theta^t\}$ are the parameters of the model ${\cal M}^t$, where $t$ represents the number of tasks considered for training the model. Let $ g_\theta(f_\omega) \colon \mathcal{X} \to \mathcal{X} $ be the encoding-decoding process for ${\cal M}$. 
	\end{definition}
	
	\begin{definition}
		\label{definition4} (\textBF{Discrepancy distance.}) We implement $h \in {\mathcal{H}}$ by $g_\theta ({f_\omega })$ evaluated on the error function $\mathcal{L} \colon \mathcal{X} \times \mathcal{X} \to \mathbb{R}_{\rm{ + }}$ which is bounded, $\forall ({\bf x},{\bf x'}) \in \mathcal{X}^2,\mathcal{L}({\bf x},{\bf x'}) \le U$ for some $U > 0$. We define the error function as the SL function $\mathcal{L}({\bf x},{\bf x'}) = \|{\bf x}-{\bf{x'}}\|^2, ({\bf x},{\bf x}') \in {\mathcal{X}}$. A risk for $h(\cdot)$ on the target distribution $\mathcal{P}_i$ of the $i$-th domain (task) is defined as ${\mathcal{R}}_{\mathcal{P}_i}(h,f_{\mathcal{P}_i}) = \mathbb{E}_{{\bf x} \sim \mathcal{P}_i} \mathcal{L} (h({\bf x}),f_{\mathcal{P}_i}({\bf x}))$, where $ f_{\mathcal{P}_i} \in \mathcal{H}$ is the true labeling function for $\mathcal{P}_i$. The discrepancy distance on two domains $\{{\mathcal{P}},{\mathbb P} \}$ over $\mathcal{X}$, is defined as:
		\begin{equation}
		\begin{aligned}
		disc_{\mathcal L}\left( {{\mathcal{P}},{\mathbb P}} \right) &= {\sup _{\left( {h,h'} \right) \in {\cal H}}}\big| {{{\mathbb E}_{{\bf{x}} \sim {\mathcal{P}} }}\left[ {{\cal L}\left( {h'\left( {\bf{x}} \right),h\left( {\bf{x}} \right)} \right)} \right] }
		\\&
		{
			- {{\mathbb E}_{{\bf{x}} \sim {\mathbb P} }}\left[ {{\cal L}\left( {h'\left( {\bf{x}} \right),h\left( {\bf{x}} \right)} \right)} \right] }\big|.
		\end{aligned}
		\label{disc}
		\end{equation}
	\end{definition}
	
	\begin{definition}
		\label{definition5} (\textBF{Empirical discrepancy distance.} ) In practice, we usually get samples of size $m_{\mathcal{P}}$ and $m_{\mathbb P}$ considering $U_{\mathcal{P}}$ and $U_{\mathbb{P}}$, respectively, and these samples form the empirical distributions $\hat{\mathcal{P}}$ and $\hat{\mathbb P}$, corresponding to ${\mathcal{P}}$ and ${\mathbb P}$. Then, the discrepancy can be estimated by using finite samples~:
		\begin{align}
		disc_\mathcal{L}\big( {{\mathcal{P}},{\mathbb P}} \big) &\le disc_{\mathcal{L}}\big( {\hat{{\mathcal{P}}},\hat{{\mathbb P}}} \big) + 8\big( {{{\mathop{\rm Re}\nolimits} }_{{U_{{\mathcal{P}} }}}}\left(\mathcal{H} \right)
		+
		{{{\mathop{\rm Re}\nolimits} }_{{U_{\mathbb P}}}}\left(\mathcal{H} \right) \big) \notag \\&+ 3M\bigg( {\sqrt {\frac{{\log \left( {\frac{4}{\delta }} \right)}}{{2{m_{{\mathcal{P}}}}}}}  + \sqrt {\frac{{\log \left( {\frac{4}{\delta }} \right)}}{{2{m_{\mathbb P}}}}} } \bigg)\,,
		\label{defition3_eq}
		\end{align}
		which holds with probability $1 - \delta,\delta \in (0,1)$, where $M>0$,  and ${{{\mathop{\rm Re}\nolimits} }_{{U_{\mathbb{P}}}}}$ is the Rademacher complexity (See Appendix-H from SM$^{1}$). We use $disc^{\star}_{\mathcal{L}}(\cdot)$ to represent the right-hand side (RHS) of Eq.~\eqref{defition3_eq}. 
	\end{definition}
	
	\section{The theoretical framework}
	\subsection{Generalization bounds for a single model}
	
	Let us consider $\mathbb{P}^i$ the approximation distribution for the generated data by $g_{\theta^i}(\cdot)$ of ${\cal M}^i$, which was trained on a sequence of domains $\{Q_1^S,\dots,Q_i^S\}$ and ${\tilde {\mathcal{P}}}_i$ represents the probabilistic representation for $Q_i^S$. Training ${\cal M}^{i+1}$ using GR, for the $(i+1)$-th task, requires the minimization of $\mathcal{L}^{\star}$ (implemented as the negative ELBO)~:
	\begin{equation}
	\begin{aligned}
	{\cal M}^{i+1} = \mathop {\arg \min }\limits_{{ \omega^{(i + 1)}},{\theta^{(i + 1)}}} \mathcal{L}^{\star}\left( {{\mathbb{P}^{i+1}},{\mathbb{P}^i} \otimes {\tilde{\mathcal{P}}_{i+1}}} \right),
	\label{optimization_fun1}
	\end{aligned}
	\end{equation}
	where $\otimes$ represents the mixing distribution $\mathbb{P}^i \otimes \tilde{\mathcal{P}}_{i+1}$, formed by samples uniformly drawn from both $\mathbb{P}^i$ and $\tilde{\mathcal{P}}_{i+1}$, respectively. Eq.~\eqref{optimization_fun1} can be treated as a recursive optimization problem as $i$ increases from 1 to $t$. The learning goal of ${\cal M}^{i+1}$ is to approximate the distribution $ \mathbb{P}^{i+1} \approx {\mathbb{P}^i} \otimes {\tilde{\mathcal{P}}_{i+1}}$ by minimizing $\mathcal{L}^{\star}(\cdot)$, when learning $(i+1)$-th task. During the LLL, the errors corresponding to the initial tasks $Q_i^S$, $i \textless t$, would increase, leading to a degenerated performance on its corresponding unseen domain, defined by its performance on the testing set $Q_i^T$. One indicator for the generalization ability of a model ${\cal M}$ is to predict its performance on a testing data set by achieving a certain error rate on a training data set \cite{TheorySourceGuided}. In this paper, we develop a new theoretical analysis that can measure the generalization of a model under LLL where the source distribution is evolved over time. Before we introduce the Generalization Bound (GB) for -ELBO, we firstly define the GB when considering a VAE learning a single task in Theorem~\ref{theorem1} and then when learning several tasks in Theorem~\ref{theorem2}.
	
	\begin{theorem}
		\label{theorem1}
		Let $\mathcal{P}_i$ and ${\tilde {\mathcal{P}}}_i$ be two domains over $\mathcal{X}$. Then for $h_{{\mathcal{P}_i}}^{*} = \arg {\min _{h \in \mathcal{H}}}{{\mathcal{R}}_{{\mathcal{P}_i}}}(h,f_{{\mathcal{P}_i}})$ and $h_{\tilde{\mathcal{P}}_i}^{*} = \arg {\min _{h \in \mathcal{H}}}{{\mathcal{R}}_{\tilde{\mathcal{P}}_i}}(h,f_{\tilde{\mathcal{P}}_i})$ where $f_{\tilde{\mathcal{P}}_i} \in \mathcal{H}$ is the ground truth function (identity function under the encoder-decoding process) for $\tilde{\mathcal{P}}_i$, we can define the GB between $\mathcal{P}_i$ and ${\tilde{\mathcal{P}}_i}$~:
		\begin{equation}
		\begin{aligned}
		{{\mathcal{R}}_{{\mathcal{P}_i}}} \big( h ,{f_{\mathcal{P}_i}} \big) &\le  \mathcal{R}_{\tilde{\mathcal{P}_i}}\big( {h,h^{*}_{\tilde{\mathcal{P}_i}}} \big) + dis{c^{\star}_{\mathcal{L}}} \big( \mathcal{P}_i,\tilde{\mathcal{P}_i} \big) \\& + {{\mathcal{R}}_{{\mathcal{P}_i}}}\big( {h_{{\mathcal{P}_i}}^{*}},{f_{\mathcal{P}_i}} \big) + {{\mathcal{R}}_{{\mathcal{P}_i}}}\big( {h_{{\mathcal{P}_i}}^{*},h^{*}_{\tilde{\mathcal{P}_i}}} \big),
		\label{theorem1_equ}
		\end{aligned}
		\end{equation}
		where the last two terms represent the optimal combined risk denoted by ${\varepsilon{({\mathcal{P}_i},{\tilde{\mathcal{P}_i}})}}$, and we have~:
		\begin{equation}
		\mathcal{R}_{\tilde{\mathcal{P}_i}}\big(h,h^{*}_{\tilde{\mathcal{P}_i}} \big) = \mathbb{E}_{{\bf x} \sim {\tilde{\mathcal{P}_i}}} \mathcal{L}\big(h({\bf{x}}),h^{*}_{\tilde{\mathcal{P}_i}}({\bf{x}})\big).
		\end{equation}
	\end{theorem}
	See the proof in Appendix-A from SM$^{1}$. We use ${\mathcal R}_A({\mathcal{P}_i},{\tilde{\mathcal{P}_i}})$ to represent $disc_{\mathcal{L}}^{\star}({\mathcal{P}_i},{\tilde{\mathcal{P}_i}})+{\varepsilon{({\mathcal{P}_i},{\tilde{\mathcal{P}_i}})}}$. Theorem~\ref{theorem1} explicitly defines the generalization error of ${\cal M}$ trained on the source distribution $\tilde{\mathcal{P}}_i$. With Theorem~\ref{theorem1}, we can extend this GB to ELBO and the marginal log-likelihood evaluation when the source distribution evolves over time.
	
	\begin{theorem}
		\label{theorem2}
		For a given sequence of tasks $\{\mathcal{T}_1,\dots,\mathcal{T}_t \}$, we derive a GB between the target distribution and the evolved source distribution during the $t$-th task learning~:
		\begin{equation}
		\begin{aligned}
		\frac{1}{t} \sum\limits_{i = 1}^t {{{\mathcal{R}}_{{\mathcal{P}_i}}}} \big( {h,{f_{{\mathcal{P}_i}}}} \big)&\le {{\mathcal{R}}_{{{\mathbb{P}}^{t - 1}} \otimes {{\tilde {\mathcal{P}}}_t}}}\big( {h,h_{{{\mathbb{P}}^{t - 1}} \otimes {{\tilde{\mathcal{P}}}_t}}^{*}} \big) \\&+ {\mathcal R}_A \big({{\mathcal{P}_{(1:t)}},{{{\mathbb{P}}^{t - 1}} \otimes {{\tilde {\mathcal{P}}}_t}}} \big),
		\label{theorem2_equ1}
		\end{aligned}
		\end{equation}
		where $\mathcal{P}_{(1:t)}$ is the mixture distribution $\{\mathcal{P}_1 \otimes \mathcal{P}_2 ,\dots,\otimes \mathcal{P}_t\}$.
	\end{theorem}
	See the proof in Appendix-B from SM$^{1}$. 
	
	\noindent \textBF{Remark.} Theorem 2 has the following observations:
	\renewcommand\labelitemi{\large$\bullet$}
	\begin{itemize}[leftmargin=10pt]
		\item The performance on the target domain depends mainly on the discrepancy term even if ${\cal M}$ minimizes the source risk, from the first term of RHS of Eq.~\eqref{theorem2_equ1}.
		\item In the GR process, ${\mathbb P}^{t-1}$ is gradually degenerated as $t$ increases due to the repeated retraining \cite{LifelongVAEGAN}, which leads to a large discrepancy distance term.
	\end{itemize}
	
	We also extend the idea from Theorem~\ref{theorem2} to derive GBs for GANs, which demonstrates that the discrepancy distance between the target and the generator's distribution plays an important role for the generalization performance of GANs under the LLL setting, exhibiting similar forgetting behaviour as VAEs (See details in Appendix-G from SM$^{1}$). In the following, we extend this GB to $\mathcal{L}^\star$.

	\begin{lemma}
		\label{lemma1}
		Let us consider the random samples 
		${\bf x}_i^T\sim{\mathcal{P}}_i$, for $i=1,\ldots,t$. The sample log-likelihood and its ELBO for all $\{ \mathcal{P}_1,\dots,\mathcal{P}_t \}$ can be represented by $\sum\nolimits_{i = 1}^t \log {p_\theta }({\bf x}_i^T)$ and $\sum\nolimits_{i = 1}^t {\mathcal{L}_{ELBO}(h,{\bf x}_i^T)}$. Let $\tilde{\bf x}^t$ represent the random sample drawn from ${\mathbb P}^{t-1}\otimes {\tilde{ \mathcal{P}}}_t$. We know that $KL(q_{\omega^t} ({\bf z}\,|\,{\bf x}_i^T)\mid\mid p({\bf{z}})) \ne {{KL}}(q_{\omega^t}({\bf{z}}\,|\,{\bf{\tilde x}}^t) \mid\mid p({\bf{z}}))$ if $q_{\omega^t}({\bf{z}}\,|\,{\bf{x}}_i^T) \ne q_{\omega^t}({\bf{z}}\,|\,{\bf{\tilde x}}^t)$, and we have~:
		\begin{equation}
		\begin{aligned}
		&\frac{1}{t}\sum\limits_{i = 1}^t {{\mathbb{E}_{{{\mathcal{P}}_i}}} KL \big(q_{\omega^t}({\bf{z}}\mid{\bf{x}}_i^T)\mid\mid p({\bf{z}}) \big) } 
		\le \\& {{\mathbb E}_{{{\mathbb P}^{t-1} \otimes {\tilde{\mathcal{P}}}_t }}}KL \big(q_{\omega^t}({\bf{z}}\mid{\bf{\tilde x}}^t)\mid\mid p({\bf{z}}) \big) + \big| KL_1 - KL_2 \big|\,,
		\label{theorem2_equ4}
		\end{aligned}
		\end{equation}
		where $q_{\omega^t} (\cdot)$ represents the inference model for ${\cal M}^t$. $KL_1$ and $KL_2$ represent the left-hand side term (LHS) and the first term of the RHS of Eq.~\eqref{theorem2_equ4}, respectively. Since ELBO consists of a negative reconstruction error term, a KL divergence term and a constant ($-\frac{1}{2}\log \pi $) \cite{Tutoria_VAEs}, when the decoder models a Gaussian distribution with a diagonal covariance matrix (the diagonal element is $1/\sqrt 2 $), we derive a GB on -ELBO by combining \eqref{theorem2_equ1} and \eqref{theorem2_equ4}~:
		\begin{align}
		&\frac{1}{t}{\sum\limits_{i = 1}^t {\mathbb{E}}_{{\mathcal{P}_i}} \Big [-{\mathcal{L}_{ELBO}\big({\bf x}^T_i ;h\big) } \Big] } \le 
		{\mathcal R}_A \big({{\mathcal{P}_{(1:t)}},{{{\mathbb{P}}^{t - 1}} \otimes {{\tilde {\mathcal{P}}}_t}}} \big) \notag
		\\&
		+\mathbb{E}_{ {\mathbb P}^{t-1}\otimes{\tilde {\mathcal{P}}}_t} \Big[-{\mathcal{L}_{ELBO}\big({\bf \tilde x}^t ;h \big) }\Big] +  \big| KL_1 - KL_2 \big| \,,
		\label{lmema1_eq1}
		\end{align}
		where ${\bf x}^T_i \sim {\mathcal{P}}_i$ and ${\tilde {\bf x}}^t \sim {\mathbb P}^{t-1}\otimes{\tilde {\mathcal{P}}}_t$. 
	\end{lemma}
	See the proof in Appendix-C from SM$^{1}$. 
	We call the RHS of Eq.~\eqref{lmema1_eq1} as Lifelong ELBO (LELBO), denoted as $\mathcal{L}_{LELBO}$ which is a bound for an infinite number of tasks ($t \to \infty $). This bound shows the behaviour of ${\cal M}$ when minimizing -ELBO when learning each task. ${\mathcal{L}}_{LELBO}$ is also an upper bound to $-\sum\nolimits_{i = 1}^t {\mathbb E}_{{\mathcal{P}}_i} \left[ \log p({\bf x}_i^T) \right] /t$, estimated by ${\cal M}^t$. 
	
	\noindent \textBF{The generalization of LELBO.} From Eq.~\eqref{lmema1_eq1}, we can generalize LELBO to other VAE variants under LLL, including the auxiliary deep generative models \cite{Aux_DGM} and hierarchical variational inference \cite{ImportanceHVAE} (See details in Appendix-F from SM$^{1}$). IWELBO bound \cite{IWVAE} is an extension of ELBO by generating multiple weighted samples under the importance sampling \cite{IWVAE2}. We generalize the IWELBO bounds to the LLL setting as:
	\begin{equation}
	\begin{aligned}
	& \frac{{\rm{1}}}{t}{\sum\limits_{i = 1}^t {\mathbb{E}}_{{\bf x}^T_i \sim {\mathcal{P}_i}} \left[ -{ \log p\left({\bf x}^T_i \right) }\right] }  \le \\
	& {\mathbb{E}}_{{\tilde{\bf x}}^t \sim {\mathbb P}^{t-1}\otimes{\tilde {\mathcal{P}}}_t} \left[  - {\mathbb{E}_{{{\bf{z}}_{1}}\dots,{\bf{z}}_{K'} \sim q\left( {{\bf{z}}\mid {\bf{x}}} \right)}}\left[ {\log \frac{1}{K’}\sum\limits_{i = 1}^{K’} {\frac{{p\left( {{{\tilde {\bf{x}}}^t},{{\bf{z}}_i}} \right)}}{{q\left( {{{\bf{z}}_i}\mid {\bf{x}}} \right)}}} } \right]
	\right] \\
	& +  \left| KL_1 - KL_2 \right| + {\mathcal R}_A \left({{\mathcal{P}_{(1:t)}},{{{\mathbb{P}}^{t - 1}} \otimes {{\tilde {\mathcal{P}}}_t}}} \right)\,.
	\label{importantce_equ4}
	\end{aligned}
	\end{equation}
	See the derivation in Appendix-F.1 from SM$^{1}$. We consider ${\bf z}_{1:K'}=\{{\bf z}_1,\dots,{\bf z}_{K'}\}$ and omit the subscript for $q(\cdot)$. $K'$ is the number of weighted samples \cite{IWVAE2}. We call RHS of Eq.~\eqref{importantce_equ4} as $\mathcal{L}_{{LELBO}_{K'}}$, and $\mathcal{L}_{{LELBO}_{K'=1 }} = \mathcal{L}_{LELBO}$.
	
	\vspace*{0.1cm}
	\noindent \textBF{Remark.} We have several conclusions from Eq.~\eqref{importantce_equ4}~:
	\vspace*{-0.1cm}
	\renewcommand\labelitemi{\large$\bullet$}
	\begin{itemize}[leftmargin=10pt]
		\item Based on the assumption that ${\mathbb P}^{t-1}$ is fixed and $|KL_1 - KL_2| = 0$, we have $\mathcal{L}_{{LELBO}_{K'+1}} \le \mathcal{L}_{{LELBO}_{K'}}$.
		\item  The tightness of ELBO on ${\mathbb P}^{t-1}\otimes{\tilde {\mathcal{P}}}_t$ (the second term of RHS of Eq.~\eqref{importantce_equ4}) can not guarantee a tight bound on the testing data log-likelihood since the RHS of Eq.~\eqref{importantce_equ4} contains the discrepancy distance term and other error terms.
	\end{itemize}
	
	A tight GB can be achieved by reducing the discrepancy distance term by training a powerful generator that approximates the target distributions well, for example by using the Autoencoding VAE \cite{AutoencodingVAE} or adversarial learning \cite{GAN}, which would fail when learning several entirely different domains due to the fixed model's capacity and the mode collapse \cite{Veegan}. In the following section, we show how we can achieve a tight GB by increasing the model's capacity through an expansion mechanism.
	
	\subsection{Generalization bounds for ENA}
	
	For a given mixture model ${\bf M} =\{{{\cal M}_1},\dots,{\cal M}_K \}$, each component ${\cal M}_i$ can be trained with GR. In order to assess the trade-off between performance and complexity, we assume that $\mathbb{P}^{(i,s)}$ is the generator distribution of the $s$-th component which was trained on a number of $i$ tasks. Suppose that the $j$-th task was learnt by the $s$-th component of the mixture and its approximation distribution $\mathbb{P}^{(m,s)}_j$ is formed by the sampling process ${\bf x}\sim\mathbb{P}^{(i,s)}$ if $I_{\mathcal{T}}({\bf x})=j$, where $I_{\mathcal{T}} \colon {\mathcal{X}} \to {\mathcal{T}}$ is the function that returns the true task label for the sample ${\bf x}$, and $m$ represents the number of times $\mathcal{M}_s$ was used with GR for the $j$-th task. We omit the component index $s$ for $\mathbb{P}^{(m,s)}_j$ for the sake of simplification and let $\mathbb{P}^0_t$ represent $\tilde{\mathcal{P}}_t$. In the following, we derive a GB for a mixture model ${\bf M}$ with $K$ components.
	
	\begin{theorem}
		\label{theorem3}
		Let $C = \{ c_1,\dots,c_m\}$ represent a set, where each item $c_i$ indicates that the $c_i$-th component (${\cal M}^1_{c_i}$) is only trained once during LLL. We use $A=\{a_1,\dots,a_m\}$ to represent the task label set for $C$, where $a_i$ is associated to $c_i$. Let $C' = \{ c'_1,\dots,c'_k\}$ represent a set where $c'_i$ indicates that the $c'_i$-th component ${\cal M}_{c'_i}$ is trained more than once and is associated with a task label set $A'_{c'_i} = \{ a{(i,1)},\dots,a{(i,n)}\} $. Let $\tilde{C}=\{ c(i,1),\dots,c(i,n) \}$ be a set where $c(i,j)$ denotes the number of times ${\cal M}_{c'_i}$ was used for $a(i,j)$-th task. We have $ |C|+|C'|=K$, $|A'_{c'_i}| > 1$, where $K$ is the number of components in the mixture model and $|\cdot|$ is the cardinality of a set. Let $\tilde A = \{ \tilde{a}_1,\dots, \tilde{a}_k\}$ represent a set where each $\tilde{a}_i$ denotes the number of tasks modelled by the probabilistic representations of the $c'_i$-th component $\tilde{a}_i=|A'_{c'_i}|$. We derive the bound for ${\bf M}$ during the $t$-th task learning~:
		\begin{equation}
		\begin{aligned}
		&\frac{1}{t} \sum\nolimits_{i = 1}^{ |C'|} \Big\{ \sum\nolimits_{j = 1}^{{{\tilde a}_i}} \Big\{ \mathcal{R}_{{\mathcal P}_{a(i,j)}}\left( {h_{c'_i},{f_{{{\mathcal P}_{a(i,j)}}}}} \right) \Big\}
		\Big\} +\\&
		\frac{1}{t}\sum\nolimits_{i = 1}^{ |C|} \Big\{  {\mathcal{R}}_{{{\mathcal P}_{a_i}}} \left( h_{c_i},f_{{\mathcal P}_{a_i}} \right) \Big\} 
		\le \frac{1}{t}{\mathcal{R}}_{C} +  \frac{1}{t}{\mathcal{R}}_{A'}
		\label{lemma3_equ1}
		\end{aligned}
		\end{equation}
		where each $h_{c_i} \in \mathcal{H}$ and $h_{c'_i} \in \mathcal{H}$ represent the hypothesis of the $c_i$-th and $c'_i$-th component in the mixture, respectively. $\mathcal{R}_C$ is the error evaluated by the components that are trained only once~:
		\begin{equation}
		\begin{aligned}
		\mathcal{R}_C &= \sum\nolimits_{i = 1}^{|C|} 
		\Big\{ \mathcal{R}_{{\tilde{\mathcal{P}}_{a_i}}} \left( h_{c_i},h^{*}_{\tilde{\mathcal{P}}_{a_i}} \right) + {\mathcal R}_A \left({{\mathcal{P}_{a_i}},{\tilde{\mathcal{P}}_{a_i}}} \right) \Big\},
		\label{RiskBoundForMixture}
		\end{aligned}
		\end{equation}
		and $\mathcal{R}_{A'}$ is the accumulated error evaluated by the components that are trained more than once~:
		\begin{equation}
		\begin{aligned}
		\mathcal{R}_{A'} &= \sum\nolimits_{i = 1}^{|C'|} \bigg\{ \sum\nolimits_{j = 1}^{\tilde{a}_i} \bigg\{ {\mathcal{R}}_{{\mathbb{P}}_{a(i,j)}^{c(i,j)}}\bigg( {h_{c'_i},{h^*_{{\mathbb{P}}_{a(i,j)}^{c(i,j)}}}} \bigg) 
		\\&+ 
		{\mathcal R}_A\bigg({{{\mathcal{P}}_{a(i,j)}},{\mathbb{P}}_{a(i,j)}^{c(i,j)}}  \bigg)
		\bigg\} \bigg\},    
		\label{Lemma3_RA_eq1}
		\end{aligned}
		\end{equation}
		and after decomposing the last term it becomes
		\begin{equation}
		\begin{aligned}
		{\mathcal R}_{A'} &= \sum\nolimits_{i = 1}^{|C'|} \bigg\{ \sum\nolimits_{j = 1}^{{{\tilde a}_i}} \bigg\{ {{\mathcal{R}}_{\mathbb{P}_{a(i,j)}^{c(i,j)}}}\bigg( {h_{c'_i},h_{{\mathbb{P}}_{a(i,j)}^{c(i,j)}}^{*}} \bigg)
		\\&{+
			\sum\nolimits_{k =  -1}^{c(i,j) - 1} {\bigg\{ {\mathcal R}_{A}\Big( {{\mathbb{P}}_{a(i,j)}^k,{\mathbb{P}}_{a(i,j)}^{k + 1}} \Big)
				\bigg\}} } \bigg\} \bigg\}\,.
		\label{theorem3_equ9}
		\end{aligned}
		\end{equation}
	\end{theorem}
	
	\noindent The proof is provided in Appendix-D from SM$^{1}$.
	\vspace*{0.1cm}
	
	\noindent \textBF{Remark.} We have several observations from \textBF{Theorem 3}~:
	\renewcommand\labelitemi{\large$\bullet$}
	\begin{itemize}[leftmargin=10pt]
		\item If $|C'| = 1$ and $|C| = 0$, then
		the term ${\mathcal R}_C$ in Eq.~\eqref{lemma3_equ1} would disappear while ${\mathcal R}_A'$ would accumulate additional error terms, according to Eq.~\eqref{theorem3_equ9}.
		\item In contrast, if $|C| = t$, then the GB from Eq.~\eqref{lemma3_equ1} is reduced to ${\mathcal R}_C$, where the number of components $K$ is equal to the number of tasks and there are no accumulated error terms, leading to a SM$^{1}$all gap on GB.
		\item When $|C|$ increases, the gap on GB tends to be small and the model's complexity tends to be large because the accumulated error term will be reduced ($|C'| = K-|C|$ in Eq.~\eqref{theorem3_equ9}) while $K$
		increases.
		\item If a single component learns multiple tasks ($|C'| = 1$), then GB on the initial tasks ($a(i,j)$ is small), tends to have more accumulated error terms compared to the GB on the latest given tasks ($a(i,j)$ is large), shown by the number of accumulated error terms $\mathcal{R}_A(\cdot,\cdot)$ in Eq.~\eqref{theorem3_equ9}, controlled by $c(i,j)=t-a(i,j)$.
	\end{itemize}
	In the following, we extend GB from $\mathcal{L}$ to $\mathcal{L}^{\star}$.
	
	\begin{lemma}
		\label{lemma4}
		We derive a GB for the marginal log-likelihood during the $t$-th task learning for $\bf{M}$:
		\begin{align}
		&\frac{1}{t} \sum\nolimits_{i = 1}^t 
		\mathbb E_{\mathcal{P}_i} \Big[ - \log p\big({\bf x}_i^T \big)\Big]
		\le \frac{1}{t} \Big( {\mathcal R}^{II}_{A'}+{\mathcal{R}}_C^{II} + D_{diff}^{\star} \Big) + \notag \\&
		\frac{1}{t} 
		\sum\nolimits_{i = 1}^{|\rm{C'}|} \bigg\{ \sum\nolimits_{j = 1}^{{\tilde{a}_i}} \bigg\{ {\mathbb E}_{{\mathbb P}^{c(i,j)}_{a(i,j)}} \Big[  - {\cal L}_{ELBO}\left( {\bf x}_{a(i,j)}^t;h_{c'_i} \right)
		\Big] \bigg\} \notag
		\\ &
		+
		\sum\nolimits_{i = 1}^{|\rm{C}|} \bigg\{
		{\mathbb E}_{{\tilde{\mathcal{P}}}_{a_i}} \Big[  - {{{\cal L}_{ELBO}}\big( {{\bf x}_{{a_i}}^S};h_{c_i} \big)}
		\Big] \bigg\} \bigg\},
		\label{Lemma2_equ1}
		\end{align}
		where we omit the component's index for each $\log p({\bf x}_i^T)$ for the sake of simplification, we use ${\mathcal R}_C^{II}$ and ${\mathcal R}_{A'}^{II}$ to represent the second terms in the RHS's from Eq.~\eqref{RiskBoundForMixture} and \eqref{Lemma3_RA_eq1}, respectively, and $D_{diff}^{\star}$ represents the absolute difference on the KL divergence (details in Appendix-E from SM$^1$). Each ${\bf x}^S_{a_i}$ is drawn from ${\tilde{\mathcal{P}}}_{a_i}$ and each ${\bf x}_{a(i,j)}^t$ is drawn from $\mathbb{P}^{c(i,j)}_{{ a}(i,j)}$ modelled by the $c'_i$-th component in ${\bf M}$. ${\mathcal{L}}_{ELBO}({\bf x}^S_{a_i} ;h_{c_i} ) $ is the ELBO estimated by the $c_i$-th component.
	\end{lemma}
	
	Lemma~\ref{lemma4} provides an explicit way to measure the gap between ELBO and the model likelihood for all tasks using the mixture model. When $|C'| = 0$, $D_{diff}^{\star} = 0$ and the discrepancy $disc_{\mathcal{L}}^{\star}({\mathcal{P}}_{a_i},\mathbb{P}_{a_i}^0)$ is very small, this bound is tight. 
	
	\section{Dynamic expansion graph model (DEGM)}
	\label{section5}
	
	According to Theorem~\ref{theorem3}, achieving an optimal GB requires each mixture component to model a unique task only. However, adding dynamically a new component whenever learning a new task, leads to ever-increasing memory and computation requirements. For addressing the trade-off between task learning effectiveness and memory efficiency, we propose a novel expansion mechanism. This would selectively allow the newly created component to reuse some of the parameters and thus transfer information from existing components, according to a knowledge similarity criterion.
	
	\subsection{Basic and specific nodes in DEGM}
	
	A component trained during LLL, with independent parameters, is called a basic node and can be transferred to be used in other tasks. Therefore, a basic node can be seen as a knowledge source for other processing nodes in DEGM.
	Meanwhile, we also have specific nodes associated with the novel information acquired from a new task ${\cal T}_{(t+1)}$, after also considering reusing the information from the basic nodes.
	
	Let $q_{\omega_i}( {\bf z} \,|\, {\bf x} )$ and $p_{\theta_i} ( {\bf x} \,|\,{\bf z})$ represent the encoding and decoding distributions, respectively, as in Eq.~\eqref{singleElbo}. We implement the basic node using paired sub-models, for encoding and decoding information. We consider two sub-inference models, $f_{\tilde{\omega_i}} \colon \mathcal{X} \to { \mathcal{\tilde Z}}$ and $f_{ {\omega'_i} } \colon { \mathcal{\tilde Z}} \to \mathcal{Z}$ for modelling $q_{\omega_i} ( {\bf z} \,|\, {\bf x} )$, expressed by $f_{\tilde{\omega_i}} \circ f_{\omega'_i} \colon \mathcal{X} \to \mathcal{Z}$, where 
	$\mathcal{\tilde Z}$ is an intermediate latent representation space with the dimension larger than $\mathcal{Z}$, $|\mathcal{\tilde Z}|>|\mathcal{Z}|$. We use two networks, $g_{\tilde {\theta_i} }\colon \mathcal{X} \to \mathcal{\tilde X}$ and $g_{ \theta'_i }\colon \mathcal{\tilde X} \to \mathcal{X}$, for modelling $p_{\theta_i}( {\bf x} \,|\, {\bf z})$ which is expressed by $g_{\tilde{\theta_i}} \circ g_{\theta'_i} \colon \mathcal{Z} \to \mathcal{X}$, where $\mathcal{\tilde X}$ is an intermediate representation space, $|\mathcal{\tilde X}|<|\mathcal{X}|$. Since a basic node has two connectable sub-models $\{f_{\tilde{\omega_i}}, g_{\tilde {\theta_i} }\}$, building a specific $j$-th node only requires two separate sub-models 
	$\{ f_{\omega'_j},g_{\theta'_j}\}$ which would be connected with the sub-models $\{f_{\tilde{\omega_i}}, g_{\tilde {\theta_i} }\}$ from all basic nodes, $i=1,\dots,K$ to form a graph structure in DEGM. In the following section, we describe how DEGM expands its architecture during LLL.
	
	\subsection{Training sub-graph structures in DEGM}
	\label{SubDMix}
	
	Let us assume that we have trained $t$ nodes after learning $t$ tasks, where $K$ nodes, $K<t$, represent basic nodes $\mathcal{G} = \{ B_1,\dots,B_K\}$, and $(t-K)$ nodes belong to specific nodes $\mathcal{S} = \{ S_1,\dots,S_{(t-K )}\}$. Let $\mathcal{GI}(\cdot)$ and $\mathcal{SI}(\cdot)$ be the functions that return the node index for ${\cal G}$ and ${\cal S}$. Each $B_i \in \mathcal{G}$ has four sub-models $\{ f_{\tilde \omega_{i^*} },f_{\omega'_{i^{*}}},g_{\tilde \theta_{i^*} },g_{\theta'_{i^{*}}}\}$ where $i^*={\mathcal{GI}}(i)$, and each $S_i \in \mathcal{S}$ has only two sub-models $\{ f_{\omega'_{i'}},g_{\theta'_{i'}}\} $, where $i'={\mathcal{SI}}(i)$. Let us consider ${\bf V} \in \mathbb{R}^{t \times t}$ an adjacency matrix representing the directed graph edges from $\mathcal{S}$ to $\mathcal{G}$. $V (i,j)$ is the directed edge from nodes $i$ to $j$, and ${\bf V}$ is used for expanding the architecture whenever necessary. After learning $t$-th task, we set a new task $\mathcal{T}_{t+1}$ for training the mixture model with $Q_{t+1}^S$. We evaluate the efficiency of using each element of $B_i \in {\cal G}$, $i=1,\ldots,K$ by calculating $\mathcal{L}_{ELBO}({\bf x}_j ;B_i )$ on ${\bf x}_j \sim Q_{t+1}^S$, $j=1,\ldots,n$, ($n=1000$ in experiments). For assessing the novelty of a given task ${\cal T}_{t+1}$, with respect to the knowledge already acquired, we consider the following criterion~:
	\begin{equation}
	\begin{aligned}
	ks_i =& \big|\mathcal{L}_{ELBO}(B_i) - {\mathbb E}_{{\bf x} \sim { Q}_{(t+1)}^S} \mathcal{L}_{ELBO}({\bf x};B_i)\big|, \;
	\label{know}
	\end{aligned}
	\end{equation}
	where $i=1,\dots,K$ and $\mathcal{L}_{ELBO}(B_i)$ is the best log-likelihood estimated by $B_i$ on its previously assigned task and we form $\mathcal{K} = \{ks_1,\dots,ks_K \}$. Similar log-likelihood evaluations were used for selecting components in \cite{NeuralDirchlet,LifelongUnsupervisedVAE}.
	However, in our approach we develop a graph-based structure by defining Basic and Specific nodes based on analyzing $\mathcal{K}$, as explained in the following.
	
	\noindent \textBF{Building a Basic node. } A Basic node is added to the DEGM model when the incoming task is assessed as completely novel. If $\min({\cal K}) > \tau $, where $\tau$ is a threshold, then we set $V(t+1,\mathcal{GI}(i) ) = 0$, $i=1,\dots,K$ and DEGM builds a basic node which is added to $\mathcal{G}$. During the $(t+1)$-th task learning, we only optimize the parameters of the $(t+1)$-th component by using the loss function from Eq.~\eqref{singleElbo} with the given task' dataset.
	
\begin{figure}[http]
		\centering
		\includegraphics[scale=0.31]{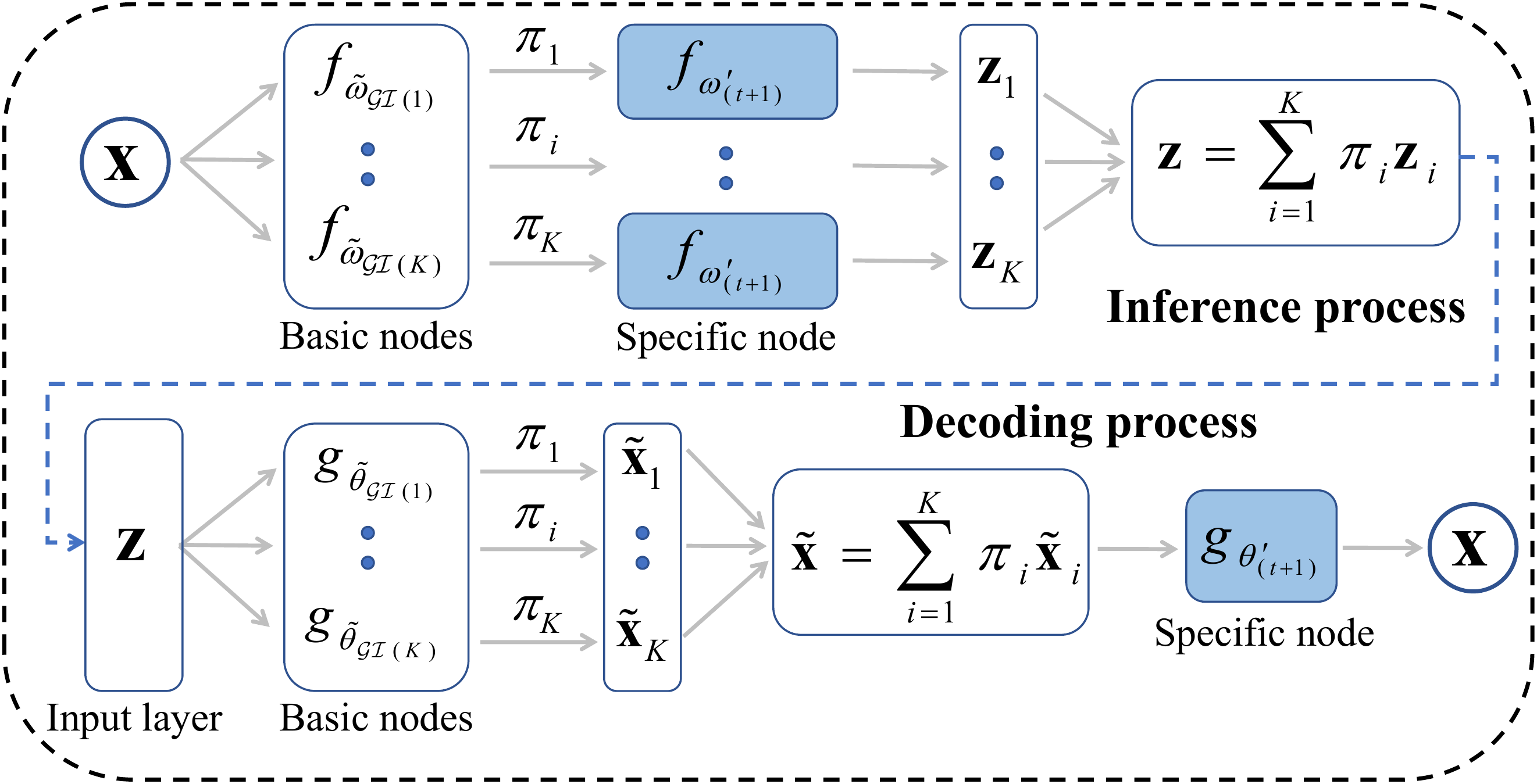}
		\caption{The graph structure in DEGM where an image is firstly processed by $K$ Basic nodes to which the newly created Specific node connects during the inference process. This procedure is also performed at the decoding process. }
		\label{grpahDMix}
	\end{figure}

	\noindent \textBF{Building a Specific node. } 
	A Specific node is built when the incoming task is related to the already learned knowledge, encoded by the basic nodes. If $\min({\cal K}) \leq \tau$, then we update ${\bf V}$ by calculating the importance weight $V(t+1,\mathcal{GI}(i) ) = (w^*-k s_i)/\sum\nolimits_{j = 1}^K (w^*- ks_j) $, $w^*=\sum\nolimits_{j = 1}^K ks_j$, $i=1,\dots,K$, where we denote $\pi_i=V(t+1,{\mathcal{GI}(i)})$ for simplification. According to the updated ${\bf V}$, we built a new sub-inference model $f_{{\omega'}_{(t+1)}}$, based on a set of sub-models $\{ f_{{{\tilde \omega }_{{i^*}}}} \,|\, i^*={\mathcal{GI}(i)},i=1,\dots,K \} $, as 
	$\sum\nolimits_{i=1}^{K }  \pi_i
	f_{{{\tilde \omega }_{i^*}}} \odot {f_{{{\omega'}_{(t + 1)}}}}({\bf x})$, which represents ${\bf z} = \sum\nolimits_{i = 1}^K \pi_i {\bf z}_i$, where each ${\bf z}_i=f_{{\tilde \omega }_{i^*}} \odot f_{{\omega '}_{(t + 1)}}({\bf x})$ is weighted by $\pi_i$. 
	In Fig.~\ref{grpahDMix}, we show the structure of the decoder, where an identity function implemented by the input layer distributes the latent variable ${\bf z}$ to each $g_{{\tilde \theta }_{i^*}}({\bf z}),i^*={\mathcal{GI}}(1),\dots,{\mathcal{GI}}(K)$, leading to $\tilde{\bf x} = \sum\nolimits_{i=1}^K \pi_i g_{{\tilde \theta }_{i^*}}({\bf z})$, where the intermediate feature information from $\mathcal{G}$ is weighted by $\pi_i$. We then build a new sub-decoder $g_{\theta'_{(t+1)}}(\tilde{\bf x})$, that takes $\tilde{\bf x}$ as the input and outputs the reconstruction of ${\bf x}$ enlarging ${\cal S}$ with $S_{t-K+1}\in \mathcal{S}$. 
	The procedure for building the graph and how a new node connects to the elements from $\mathcal{G}$, as a sub-graph in DEGM, is shown in Fig.~\ref{grpahDMix}. The importance of processing modules during LLL was considered in \cite{TaskFreeCL,Lifelong_NodeImportance}. However, DEGM is the first model where this mechanism is used for the dynamic expansion of a graph model. Additionally, different from existing methods, the importance weighting approach proposed in this paper regularizes the transferable information during both the inference and generation processes. In the following, we propose a new objective function for the training of a Specific node, which also guarantees a lower bound to the marginal log-likelihood.
	
	\begin{theorem}
		\label{theo4}
		A Specific node is built for learning the $(t+1)$-th task, which forms a sub-graph structure and can be trained by using a valid lower bound (ELBO) (See details in Appendix-J.1 from SM$^1$)~:
		\begin{equation}
		\begin{aligned}
		&\mathcal{L}_{MELBO}({\bf x};{\cal M}_{(t+1)}) =: \\& {\mathbb{E}_{Q\left(\bf z \right)}} \left[ \log {p_{{\theta'_{(t + 1)}} \circ \{ {\tilde \theta}_{\mathcal{GI}(1)},\dots,{\tilde \theta}_{\mathcal{GI}(K)}\}
		}}( {\bf x} \mid {\bf z}) \right] \\
		&- \sum\nolimits_{i = 1}^K
		\pi_i{{KL}\left( {Q_{{{{\tilde \omega }_{\mathcal{GI}(i)}}} \, \circ \, {\omega '}_{(t + 1)} } \left( {{\bf{z}}\mid{\bf x}} \right)} \mid\mid p\left( {\bf z}_i \right) \right)}\,,
		\label{MixtureElbo}
		\end{aligned}
		\end{equation}
		where ${q_{{{\tilde \omega }_{\mathcal{GI}(i)}}}} \circ {q_{{{\omega '}_{(t + 1)}}}}\left( {{\bf{z}} \,|\, {\bf{x}}} \right)$ is the density function form of ${Q_{{{{\tilde \omega }_{\mathcal{GI}(i)}}} \, \circ \, {\omega '}_{(t + 1)} } \left( {\bf z} \,|\, {\bf x} \right)} $.
	\end{theorem}
	We implement the variational distribution $Q({\bf z})$ by $\sum\nolimits_{i = 1}^K \pi_i{Q_{{{{\tilde \omega }_{\mathcal{GI}(i)}}} \, \circ \, {\omega '}_{(t + 1)} } \left( {\bf z}\,|\,{\bf x} \right)}$, which is a mixture inference model.
	The difference in Eq.~\eqref{MixtureElbo} from $q_\omega({\bf z} \,|\, {\bf x})$ in Eq.~\eqref{singleElbo} is that $Q(\bf z)$ is much more knowledge expressive by reusing the transferable information from previously learnt knowledge while also reducing the computational cost, by only updating the components $\{ \omega'_{(t + 1)},\theta'_{(t + 1)}\}$ when learning the $(t+1)$-th task. The first term in the RHS of Eq.~\eqref{MixtureElbo} is the negative reconstruction error evaluated by a single decoder. The second term consists of the sum of all KL terms weighted by their corresponding edge values $\pi_i$.\\
	\noindent \textBF{Model selection.} We evaluate the negative marginal log-likelihood, (Eq.~\eqref{singleElbo} for Basic nodes, and Eq.~\eqref{MixtureElbo} for Specific nodes) for testing data samples after LLL. Then we choose the node with the highest likelihood for the evaluation. This mechanism can allow DEGM to infer an appropriate node without task labels (See details in Appendix-J.3 from SM$^1$).
	
	\section{Experimental results}
	
	\subsection{Unsupervised lifelong learning benchmark}
	
	\textBF{Setting.} We define a novel benchmark for the log-likelihood estimation under LLL, explained in Appendix-K from SM$^1$. We consider learning multiple tasks defined within a single domain, such as MNIST \cite{MNIST} and Fashion \cite{FashionMNIST}. Following from \cite{IWVAE} we divide MNIST and Fashion into five tasks \cite{Continual_Learning}, called Split MNIST (S-M) and 
	Split- Fashion (S-F). We use the cross-domain setting where we aim to learn a sequence of domains, called COFMI, consisting of: Caltech 101 \cite{Caltech101}, OMNIGLOT \cite{Omniglot}, Fashion, MNIST, InverseFashion (IFashion) where each task is associated with a distinct dataset from all others. All databases are binarized.
	
	\noindent \textBF{Baselines.} We adapt the network architecture from \cite{IWVAE} and consider several baselines. A single VAE with GR is called ELBO-GR and when considering IWELBO bounds it becomes IWELBO-GR-$K'$ where $K'$ represent the number of weighted samples. We call DEGM with ELBO and IWELBO bounds as DEGM-ELBO and DEGM-IWELBO-$K'$, respectively. We also compare with LIMix \cite{LifelongInfinite} and implement CN-DPM \cite{NeuralDirchlet} with the optimal setting, namely CN-DPM* (See details in Appendix-L.1 from SM$^1$).
	
	\renewcommand\arraystretch{1.1}
	\begin{table}[h]
		\small
		\centering
		\setlength{\tabcolsep}{3.2mm}{
			\begin{tabular}{@{}l cc cc@{} } 
				\toprule 
				\textbf{Methods}   & S-M & S-F & COFMI \\
				\midrule 
				ELBO-GR& -98.23&-240.58 & -177.47  \\
				IWELBO-GR-50&-93.57 &-236.66&-172.10 \\
				IWELBO-GR-5&-95.80 &-238.08&-176.21  \\
				ELBO-GR*& -98.36&-243.91& -180.50  \\
				IWELBO-GR*-50&-91.23&-236.90&-188.9\\
				CN-DPM*-IWELBO-50&-95.91&-237.47&-184.19 \\
				LIMix-IWELBO-50 &-95.74&-237.48&-184.32\\
				\midrule 
				\midrule 
				DEGM-ELBO &-93.51&-238.54&-168.91  \\
				DEGM-IWELBO-50 &\textBF{-88.04} & \textBF{ -233.76} &\textBF{-163.27}\\
				DEGM-IWELBO-5 &-91.44&-235.93&-164.99  \\
				\bottomrule 
			\end{tabular}
			\caption{Results for Split MNIST, Split Fashion and COFMI.}
			\label{basic_tab}
		}
	\end{table}
	
	\noindent \textBF{Results.} The testing data log-likelihood is estimated by the IWELBO bounds \cite{IWVAE} with $K'=5000$. we perform five independent runs for S-M/S-F and COFMI data. The average results are reported in Table~\ref{basic_tab} where `*' denotes that the model uses two stochastic layers (See details in Appendix-K.1 from SM$^1$) which can further improve the performance with the IWELBO bound, according to IWELBO-GR*-50. The proposed DEGM-IWELBO-50 obtains the best results when using the IWELBO bound, for both S-M and S-F settings. The proposed DEGM also outperforms other baselines on COFMI, which represents a more challenging task than S-M/S-F. The detailed results for each task are reported in Appendix-K.2 from SM$^1$, showing that ELBO-GR* and IWELBO-GR*-50 tend to degenerate the performance on the early tasks under the cross-domain learning setting when compared with VAEs that do not use two stochastic layers. Details, such as the number of Basic and Specific nodes used are provided in Appendix-K.2 from SM$^1$.
	
	\vspace*{-0.1cm}
	\subsection{Comparing to lifelong learning models}
	\vspace*{0.1cm}
	\noindent \textBF{Baselines.} The first baseline consists of dynamically creating a new VAE to adapt to a new task, namely DEGM-2, which is a strong baseline and would achieve the best performance for each new task. Meanwhile, Batch Ensemble (BE) \cite{BatchEnsemble} is designed for classification tasks. We implement each component of BE as a VAE. DEGM is trained by using ELBO without considering the IWELBO bound aiming for using a small network. The number of parameters required by various models is listed in Appendix-L.5 from SM$^1$. 
	
	We train various models under CCCOSCZC lifelong learning setting, where each task is associated with a dataset: CelebA \cite{Celeba}, CACD \cite{CACD}, 3D-Chair \cite{3DChairs}, Ommiglot \cite{Omniglot}, ImageNet* \cite{ImageNet}, Car \cite{CompCars}, Zappos \cite{shore_Dataset}, CUB \cite{CUB_Birds}  (detailed dataset setting is provided in Appendix-L.1 from SM$^1$). The square loss (SL) is used to evaluate the reconstruction quality and other criteria are reported in Appendix-L.3 from SM$^1$. The threshold for adding a new component for DEGM is $\tau=600$  on CCCOSCZC and the results are reported in Table~\ref{ComplicatedTasks}. We can observe that the proposed DEGM outperforms other existing lifelong learning models and achieves a close result to DEGM-2 which trains individual VAEs for each task and requires more parameters. Visual results are shown in Appendix-L.7 from SM$^1$.
	
	\begin{table}[http]
		\centering
		\small
		\setlength{\tabcolsep}{0.75mm}{
			\begin{tabular}{@{} lcccccc @{}}
				\toprule
				Dataset & BE&LIMix  & LGM &DEGM & DEGM-2 &CN-DPM* \\
				\midrule
				CelebA &213.9&214.2& 535.6&229.2&217.0&215.4
				\\
				CACD &414.9&353.5& 814.3&368.3&281.95&347.3
				\\
				3D-Chair &649.1&353.1&
				2705.9&324.0&291.46&513.8
				\\
				Omniglot &875.1&351.1&
				5958.9&225.6&195.7&343.2
				\\
				ImageNet* &758.4&778.5&
				683.1&689.6&652.8&769.1
				\\
				Car &745.1&688.19&
				583.7&588.8&565.9&709.8
				\\
				Zappos &451.1&283.4&
				431.2&263.4&275.8&280.7
				\\
				CUB &492.0&400.7&
				330.2&461.3&569.6&638.6
				\\
				Average &575.0&427.8&
				1505.4&393.8&381.3&477.2
				\\
				\bottomrule
			\end{tabular}
			\caption{Results under CCCOSCZC lifelong learning.}
			\label{ComplicatedTasks}
		}
	\end{table}
	
	\vspace*{-0.2cm}
	\subsection{Empirical results for theoretical analysis}
	\vspace*{0.2cm}
	
\begin{figure}[h]
		\hspace{-8pt}
		\centering
		\subfigure[Target risk as in Eq.~\eqref{importantce_equ4}.]{
			\centering
			\includegraphics[scale=0.285]{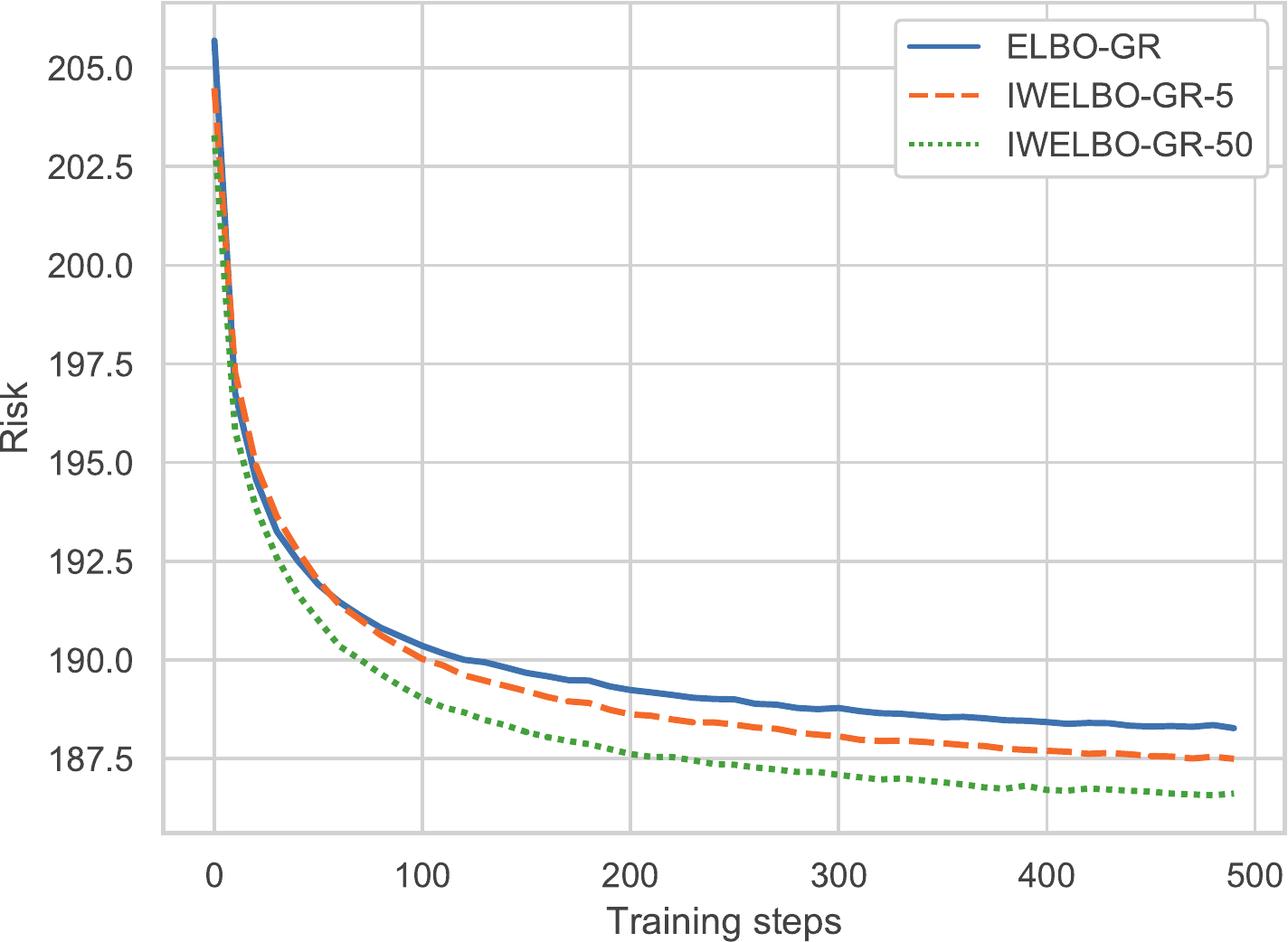}
		}
		\hspace{-5pt}
		\subfigure[Evaluation of Eq.~\eqref{lmema1_eq1}.]{
			\centering
			\includegraphics[scale=0.285]{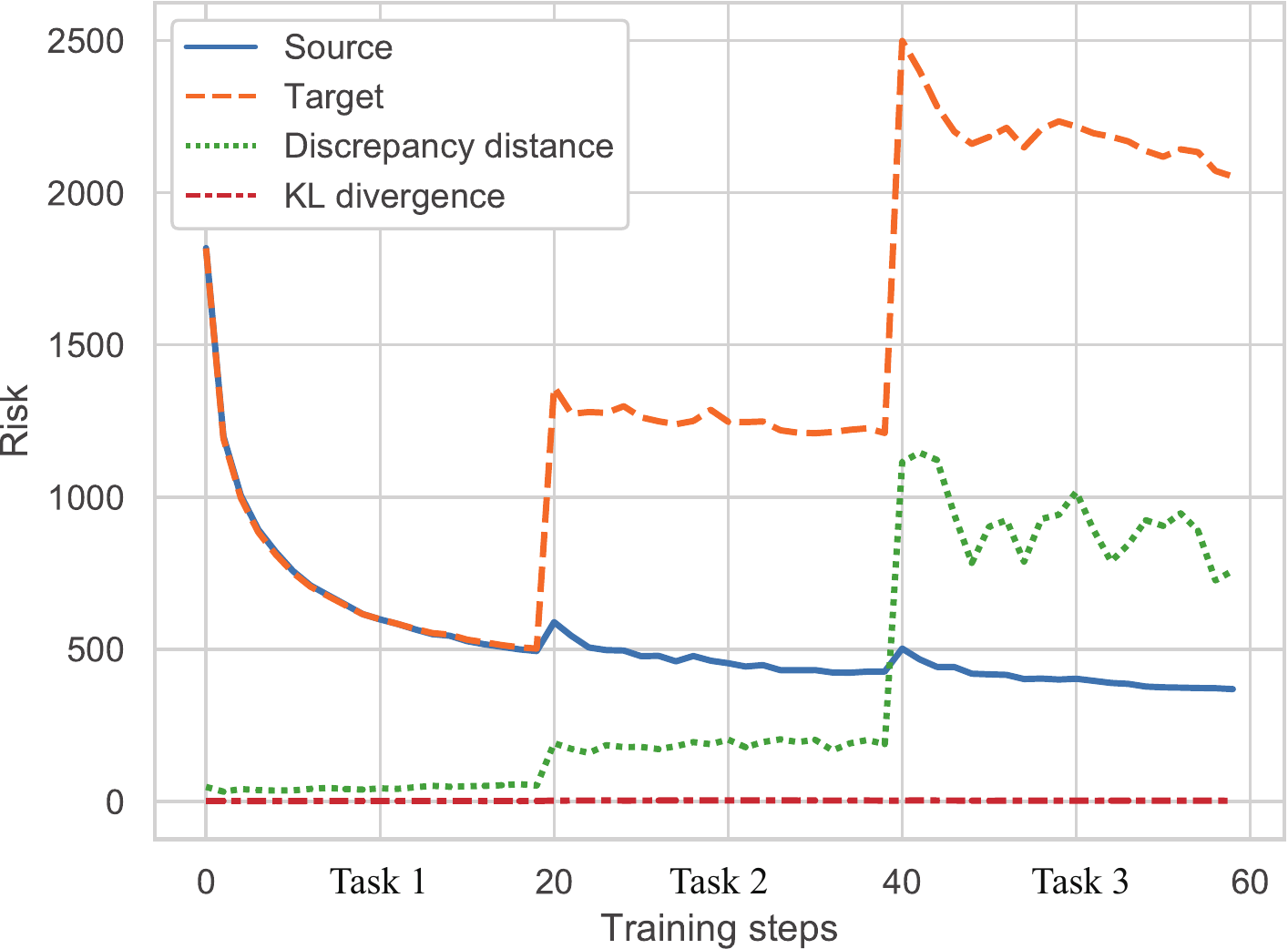}
		}
		\vspace{-8pt}
		\caption{The estimation of the target and source risks.}
		\label{analysis_results}
	\end{figure}

	We train a VAE on the binarized Caltech 101 database and use it to generate a pseudo set of images consistent with the dataset. Then, ELBO-GR, IWELBO-GR-$K'$, where $K' \in \{ 5, 50 \}$ corresponds to the number of weighted samples used for training on the joint dataset consisting of the pseudo set and a training set from a second task (Fashion). We evaluate the average target risk (LHS of Eq.~\eqref{importantce_equ4}) for these models in order to investigate the tightness between the negative log-likelihood (NLL) and LELBO, since NLL is a lower bound to LELBO. The IWELBO bounds with 5000 weighted samples results are shown in Fig.~\ref{analysis_results}a. Although, Lemma~\ref{lemma1} considers the Gaussian decoder, VAEs with a Bernoulli decoder, corresponding to the IWELBO bound, indicates that IWELBO-GR-50 is a lower bound to IWELBO-GR-5 and ELBO-GR, which empirically proves $\mathcal{L}_{{LELBO}_{50}} \le \mathcal{L}_{{LELBO}_{5}}$ when the generator distribution is fixed and $|KL_1 - KL_2| = 0$, as discussed in Lemma~\ref{lemma1}. 
	
	We also train a VAE whose decoder outputs the mean vector of a Gaussian distribution with the Identity matrix as its covariance, under MNIST, Fashion and IFashion LLL, where pixel values of all images are within $(0,255)$. The reconstruction error ELBO is normalized by dividing with the image size ($28\times28$), as in \cite{VLVAE}. We evaluate the risk and the discrepancy distance for each training epoch, according to Eq.~\eqref{lmema1_eq1} from Lemma~\ref{lemma1} and the results are provided in Fig.~\ref{analysis_results}b, where the source risk (the first term in RHS of Eq.~\eqref{lmema1_eq1}) keeps stable and the discrepancy distance ${disc}_{\mathcal{L}}(\cdot)$, Eq.~\eqref{disc}, represented within ${\mathcal{R}}_A(\cdot)$, increases while learning more tasks. The `KL divergence,' calculated as $|KL_1 - KL_2|$, shown in Fig.~\ref{analysis_results}b increases slowly. This demonstrates that the discrepancy distance plays an important role on shrinking the gap for the GB. An ablation study, demonstrating the effectiveness of the proposed expansion mechanism, is provided in Appendix-L.4 from SM$^1$. 
	
	\section{Conclusion}
	
	In this paper we analyze the forgetting behaviour of VAEs by finding an upper bound on the negative marginal log-likelihood, called LELBO. This provides insights into the generalization performance on the target distribution when the source distribution evolves continuously over time during lifelong learning. We further develop a Dynamic Expansion Graph Model (DEGM), which adds new Basic and Specific components to the network, depending on a knowledge novelty criterion. DEGM can significantly reduce the accumulated errors caused by the forgetting process. The empirical and theoretical results verify the effectiveness of the proposed DEGM methodology.

	{
		\bibliography{egbib}

\begin{thebibliography}{59}
\providecommand{\natexlab}[1]{#1}

\bibitem[{Achille et~al.(2018)Achille, Eccles, Matthey, Burgess, Watters,
  Lerchner, and Higgins}]{Lifelong_VAE}
Achille, A.; Eccles, T.; Matthey, L.; Burgess, C.; Watters, N.; Lerchner, A.;
  and Higgins, I. 2018.
\newblock Life-long disentangled representation learning with cross-domain
  latent homologies.
\newblock In \emph{Advances in Neural Inf. Proc. Systems (NIPS)}, 9873--9883.

\bibitem[{Aljundi, Kelchtermans, and Tuytelaars(2019)}]{TaskFreeCL}
Aljundi, R.; Kelchtermans, K.; and Tuytelaars, T. 2019.
\newblock Task-free continual learning.
\newblock In \emph{Proc. of the IEEE/CVF Conference on Computer Vision and
  Pattern Recognition}, 11254--11263.

\bibitem[{Aubry et~al.(2014)Aubry, Maturana, Efros, Russell, and
  Sivic}]{3DChairs}
Aubry, M.; Maturana, D.; Efros, A.~A.; Russell, B.~C.; and Sivic, J. 2014.
\newblock Seeing {3D} chairs: exemplar part-based 2d-3d alignment using a large
  dataset of cad models.
\newblock In \emph{Proc. of IEEE Conf. on Computer Vision and Pattern
  Recognition (CVPR)}, 3762--3769.

\bibitem[{Burda, Grosse, and Salakhutdinov(2015)}]{IWVAE}
Burda, Y.; Grosse, R.; and Salakhutdinov, R. 2015.
\newblock Importance weighted autoencoders.
\newblock \emph{arXiv preprint arXiv:1509.00519}.

\bibitem[{Cemgil et~al.(2020)Cemgil, Ghaisas, Dvijotham, Gowal, and
  Kohli}]{AutoencodingVAE}
Cemgil, T.; Ghaisas, S.; Dvijotham, K.; Gowal, S.; and Kohli, P. 2020.
\newblock The Autoencoding Variational Autoencoder.
\newblock In \emph{Advances in Neural Information Processing Systems
  (NeurIPS)}, volume~33, 15077--15087.

\bibitem[{Chaudhry et~al.(2018)Chaudhry, Ranzato, Rohrbach, and
  Elhoseiny}]{AGEM}
Chaudhry, A.; Ranzato, M.; Rohrbach, M.; and Elhoseiny, M. 2018.
\newblock Efficient lifelong learning with {A-GEM}.
\newblock In \emph{Proc. Int. Conf. on Learning Representations (ICLR), arXiv
  preprint arXiv:1812.00420}.

\bibitem[{Chen, Chen, and Hsu(2014)}]{CACD}
Chen, B.-C.; Chen, C.-S.; and Hsu, W.~H. 2014.
\newblock Cross-Age Reference Coding for Age-Invariant Face Recognition and
  Retrieval.
\newblock In \emph{Proc. European Conf on Computer Vision (ECCV), vol. LNCS
  8694}, 768--783.

\bibitem[{Chen et~al.(2018)Chen, Dai, Pu, Li, Su, and Carin}]{VAE_symmetric}
Chen, L.; Dai, S.; Pu, Y.; Li, C.; Su, Q.; and Carin, L. 2018.
\newblock Symmetric variational autoencoder and connections to adversarial
  learning.
\newblock In \emph{Proc. Int. Conf. on Artificial Intel. and Statistics
  (AISTATS) 2018, vol. PMLR 84}, 661--669.

\bibitem[{Doersch(2016)}]{Tutoria_VAEs}
Doersch, C. 2016.
\newblock Tutorial on variational autoencoders.
\newblock \emph{arXiv preprint arXiv:1606.05908}.

\bibitem[{Domke and Sheldon(2018)}]{IWVAE2}
Domke, J.; and Sheldon, D.~R. 2018.
\newblock Importance weighting and variational inference.
\newblock In \emph{Advances in Neural Information Processing Systems
  (NeurIPS)}, 4470--4479.

\bibitem[{Fei-Fei, Fergus, and Perona(2004)}]{Caltech101}
Fei-Fei, L.; Fergus, R.; and Perona, P. 2004.
\newblock Learning generative visual models from few training examples: An
  incremental bayesian approach tested on 101 object categories.
\newblock In \emph{IEEE CVPR-workshop}, 1--9.

\bibitem[{French(1999)}]{Catastrophic}
French, R.~M. 1999.
\newblock Catastrophic forgetting in connectionist networks.
\newblock \emph{Trends in cognitive sciences}, 3(4): 128--135.

\bibitem[{Goodfellow et~al.(2014)Goodfellow, Pouget-Abadie, Mirza, Xu,
  Warde-Farley, Ozair, Courville, and Bengio}]{GAN}
Goodfellow, I.~J.; Pouget-Abadie, J.; Mirza, M.; Xu, B.; Warde-Farley, D.;
  Ozair, S.; Courville, A.; and Bengio, Y. 2014.
\newblock Generative adversarial nets.
\newblock In \emph{Advances in Neural Inf. Proc. Systems (NIPS)}, 2672--2680.

\bibitem[{Guo et~al.(2020)Guo, Liu, Yang, and Rosing}]{ImprovedAGEM}
Guo, Y.; Liu, M.; Yang, T.; and Rosing, T. 2020.
\newblock Improved Schemes for Episodic Memory-based Lifelong Learning.
\newblock In \emph{Advances in Neural Information Processing Systems
  (NeurIPS)}, volume~33, 1--13.

\bibitem[{Higgins et~al.(2017)Higgins, Matthey, Pal, Burgess, Glorot,
  Botvinick, Mohamed, and Lerchner}]{baeVAE}
Higgins, I.; Matthey, L.; Pal, A.; Burgess, C.; Glorot, X.; Botvinick, M.;
  Mohamed, S.; and Lerchner, A. 2017.
\newblock $\beta$-{VAE}: Learning basic visual concepts with a constrained
  variational framework.
\newblock In \emph{Proc. Int. Conf. on Learning Representations (ICLR)}, 1--13.

\bibitem[{Huang et~al.(2019)Huang, Sankaran, Dhekane, Lacoste, and
  Courville}]{HImportedVAE}
Huang, C.-W.; Sankaran, K.; Dhekane, E.; Lacoste, A.; and Courville, A. 2019.
\newblock Hierarchical importance weighted autoencoders.
\newblock In \emph{Int. Conf. on Machine Learning (ICML), vol. PMLR 97},
  2869--2878.

\bibitem[{Jung, Jung, and Kim(2016)}]{LessForgetting}
Jung, H.; Jung, M.; and Kim, J. 2016.
\newblock Less-forgetting learning in deep neural networks.
\newblock \emph{arXiv preprint arXiv:1607.00122}.

\bibitem[{Jung et~al.(2020)Jung, Ahn, Cha, and Moon}]{Lifelong_NodeImportance}
Jung, S.; Ahn, H.; Cha, S.; and Moon, T. 2020.
\newblock Continual Learning with Node-Importance based Adaptive Group Sparse
  Regularization.
\newblock In \emph{Advances in Neural Information Processing Systems
  (NeurIPS)}, 1--12.

\bibitem[{Kim and Pavlovic(2020)}]{RecursiveVAE}
Kim, M.; and Pavlovic, V. 2020.
\newblock Recursive Inference for Variational Autoencoders.
\newblock In \emph{Advances in Neural Information Processing Systems},
  volume~33, 19632--19641.

\bibitem[{Kingma et~al.(2016)Kingma, Salimans, Jozefowicz, Chen, Sutskever, and
  Welling}]{ImpVarInf}
Kingma, D.~P.; Salimans, T.; Jozefowicz, R.; Chen, X.; Sutskever, J.; and
  Welling, M. 2016.
\newblock Improved variational inference with inverse autoregressive flow.
\newblock In \emph{Proc. Advances in Neural Inf. Proc. Systems (NIPS)},
  4743--4751.

\bibitem[{Kingma and Welling(2013)}]{VAE}
Kingma, D.~P.; and Welling, M. 2013.
\newblock Auto-encoding variational {B}ayes.
\newblock \emph{arXiv preprint arXiv:1312.6114}.

\bibitem[{Krizhevsky, Sutskever, and Hinton(2012)}]{ImageNet}
Krizhevsky, A.; Sutskever, I.; and Hinton, G.~E. 2012.
\newblock Imagenet classification with deep convolutional neural networks.
\newblock In \emph{Proc. Advances in Neural Inf. Proc. Systems (NIPS)},
  1097--1105.

\bibitem[{Kuroki et~al.(2019)Kuroki, Charoenphakdee, Bao, Honda, Sato, and
  Sugiyama}]{TheorySourceGuided}
Kuroki, S.; Charoenphakdee, N.; Bao, H.; Honda, J.; Sato, I.; and Sugiyama, M.
  2019.
\newblock Unsupervised domain adaptation based on source-guided discrepancy.
\newblock In \emph{Proc. AAAI Conf. on Artificial Intelligence}, volume~33,
  4122--4129.

\bibitem[{Lake, Salakhutdinov, and Tenenbaum(2015)}]{Omniglot}
Lake, B.~M.; Salakhutdinov, R.; and Tenenbaum, J.~B. 2015.
\newblock Human-level concept learning through probabilistic program induction.
\newblock \emph{Science}, 350(6266): 1332--1338.

\bibitem[{LeCun et~al.(1998)LeCun, Bottou, Bengio, and Haffner}]{MNIST}
LeCun, Y.; Bottou, L.; Bengio, Y.; and Haffner, P. 1998.
\newblock Gradient-based learning applied to document recognition.
\newblock \emph{Proc. of the IEEE}, 86(11): 2278--2324.

\bibitem[{Lee et~al.(2020)Lee, Ha, Zhang, and Kim}]{NeuralDirchlet}
Lee, S.; Ha, J.; Zhang, D.; and Kim, G. 2020.
\newblock A Neural {D}irichlet Process Mixture Model for Task-Free Continual
  Learning.
\newblock In \emph{Proc. Int. Conf. on Learning Representations (ICLR), arXiv
  preprint arXiv:2001.00689}.

\bibitem[{Li and Hoiem(2017)}]{Lwf}
Li, Z.; and Hoiem, D. 2017.
\newblock Learning without forgetting.
\newblock \emph{IEEE Trans. on Pattern Analysis and Machine Intelligence},
  40(12): 2935--2947.

\bibitem[{Liu et~al.(2015)Liu, Luo, Wang, and Tang}]{Celeba}
Liu, Z.; Luo, P.; Wang, X.; and Tang, X. 2015.
\newblock Deep learning face attributes in the wild.
\newblock In \emph{Proc. of IEEE Int. Conf. on Computer Vision (ICCV)},
  3730--3738.

\bibitem[{Maal$\o$e et~al.(2016)Maal$\o$e, S$\o$nderby, S$\o$nderby, and
  Winther}]{Aux_DGM}
Maal$\o$e, L.; S$\o$nderby, C.~K.; S$\o$nderby, S.~K.; and Winther, O. 2016.
\newblock Auxiliary deep generative models.
\newblock In \emph{Proc. Int. Conf. on Machine Learning (ICML), vol. PMLR 48},
  1445--1453.

\bibitem[{Mansour, Mohri, and Rostamizadeh(2009)}]{domainTheory}
Mansour, Y.; Mohri, M.; and Rostamizadeh, A. 2009.
\newblock Domain adaptation: Learning bounds and algorithms.
\newblock In \emph{Proc. Conf. on Learning Theory (COLT), arXiv preprint
  arXiv:2002.06715}.

\bibitem[{Mescheder, Nowozin, and Geiger(2017)}]{AdbVB}
Mescheder, L.; Nowozin, S.; and Geiger, A. 2017.
\newblock Adversarial Variational {Bayes}: Unifying variational autoencoders
  and generative adversarial networks.
\newblock In \emph{Proc. Int. Conf. on Machine Learning (ICML), vol. PMLR 70},
  2391--2400.

\bibitem[{Molchanov et~al.(2019)Molchanov, Kharitonov, Sobolev, and
  Vetrov}]{DoublySemiVAE}
Molchanov, D.; Kharitonov, V.; Sobolev, A.; and Vetrov, D. 2019.
\newblock Doubly semi-implicit variational inference.
\newblock In \emph{Proc. Int. Conf. on Artificial Intelligence and Statistics
  (AISTATS), vol. PMLR 89}, 2593--2602.

\bibitem[{Nguyen et~al.(2017)Nguyen, Li, Bui, and Turner}]{VCL}
Nguyen, C.~V.; Li, Y.; Bui, T.~D.; and Turner, R.~E. 2017.
\newblock Variational continual learning.
\newblock In \emph{Proc. Int. Conf. on Learning Representations (ICLR), arXiv
  preprint arXiv:1710.10628}.

\bibitem[{Pan et~al.(2020)Pan, Swaroop, Immer, Eschenhagen, Turner, and
  Khan}]{Functional_Regularisation_LLL}
Pan, P.; Swaroop, S.; Immer, A.; Eschenhagen, R.; Turner, R.; and Khan, M.
  E.~E. 2020.
\newblock Continual Deep Learning by Functional Regularisation of Memorable
  Past.
\newblock In \emph{Advances in Neural Information Processing Systems
  (NeurIPS)}, volume~33, 4453--4464.

\bibitem[{Park, Kim, and Kim(2019)}]{VLVAE}
Park, Y.; Kim, C.; and Kim, G. 2019.
\newblock Variational {L}aplace autoencoders.
\newblock In \emph{Proc. Int. Conf. on Machine Learning (ICML), vol. PMLR 97},
  5032--5041.

\bibitem[{Ramapuram, Gregorova, and Kalousis(2020)}]{GenerativeLifelong}
Ramapuram, J.; Gregorova, M.; and Kalousis, A. 2020.
\newblock Lifelong Generative Modeling.
\newblock \emph{Neurocomputing}, 404: 381--400.

\bibitem[{Rao et~al.(2019)Rao, Visin, Rusu, Teh, Pascanu, and
  Hadsell}]{LifelongUnsupervisedVAE}
Rao, D.; Visin, F.; Rusu, A.~A.; Teh, Y.~W.; Pascanu, R.; and Hadsell, R. 2019.
\newblock Continual Unsupervised Representation Learning.
\newblock In \emph{Advances in Neural Information Processing Systems
  (NeurIPS)}, 1--11.

\bibitem[{Rezende and Mohamed(2015)}]{VAE_NormFlow}
Rezende, D.~J.; and Mohamed, S. 2015.
\newblock Variational inference with normalizing flows.
\newblock In \emph{Proc. Int. Conf. on Machine Learning (ICML), vol. PMLR 37},
  1530--1538.

\bibitem[{Riemer et~al.(2019)Riemer, Cases, Ajemian, Liu, Rish, Tu, , and
  Tesauro}]{InterferenceLifelongLearning}
Riemer, M.; Cases, I.; Ajemian, R.; Liu, M.; Rish, I.; Tu, Y.; ; and Tesauro,
  G. 2019.
\newblock Learning to Learn without Forgetting By Maximizing Transfer and
  Minimizing Interference.
\newblock In \emph{Proc. Int. Conf. on Learning Representations (ICLR), arXiv
  preprint arXiv:1810.11910}.

\bibitem[{Shin et~al.(2017)Shin, Lee, Kim, and Kim}]{Generative_replay}
Shin, H.; Lee, J.~K.; Kim, J.; and Kim, J. 2017.
\newblock Continual learning with deep generative replay.
\newblock In \emph{Advances in Neural Information Proc. Systems (NIPS)},
  2990--2999.

\bibitem[{Sobolev and Vetrov(2019)}]{ImportanceHVAE}
Sobolev, A.; and Vetrov, D. 2019.
\newblock Importance Weighted Hierarchical Variational Inference.
\newblock In \emph{Advances in Neural Information Processing Systems
  (NeurIPS)}, volume~32, 1--13.

\bibitem[{Srivastava et~al.(2017)Srivastava, Valkov, Russell, Gutmann, and
  Sutton}]{Veegan}
Srivastava, A.; Valkov, L.; Russell, C.; Gutmann, M.~U.; and Sutton, C. 2017.
\newblock {VEEGAN}: Reducing mode collapse in GANs using implicit variational
  learning.
\newblock In \emph{Advances in Neural Inf. Proc. Systems (NIPS)}, 3308--3318.

\bibitem[{Vahdat and Kautz(2020)}]{NVAE}
Vahdat, A.; and Kautz, J. 2020.
\newblock {NVAE}: A Deep Hierarchical Variational Autoencoder.
\newblock In \emph{Advances in Neural Information Processing Systems
  (NeurIPS)}, volume~33, 19667--19679.

\bibitem[{Wah et~al.(2010)Wah, Branson, Welinder, Perona, and
  Belongie}]{CUB_Birds}
Wah, C.; Branson, S.; Welinder, P.; Perona, P.; and Belongie, S. 2010.
\newblock The {C}altech-{UCSD} Birds-200 dataset.
\newblock Technical Report CNS-TR-2010-001, California Institute of Technology.

\bibitem[{Wen, Tran, and Ba(2020)}]{BatchEnsemble}
Wen, Y.; Tran, D.; and Ba, J. 2020.
\newblock {BatchEnsemble}: an Alternative Approach to Efficient Ensemble and
  Lifelong Learning.
\newblock In \emph{Proc. Int. Conf. on Learning Representations (ICLR), arXiv
  preprint arXiv:2002.06715}.

\bibitem[{Xiao, Rasul, and Vollgraf(2017)}]{FashionMNIST}
Xiao, H.; Rasul, K.; and Vollgraf, R. 2017.
\newblock Fashion-{MNIST}: a novel image dataset for benchmarking machine
  learning algorithms.
\newblock \emph{arXiv preprint arXiv:1708.07747}.

\bibitem[{Yang et~al.(2015)Yang, Luo, Change~Loy, and Tang}]{CompCars}
Yang, L.; Luo, P.; Change~Loy, C.; and Tang, X. 2015.
\newblock A large-scale car dataset for fine-grained categorization and
  verification.
\newblock In \emph{Proc. IEEE Conf. on Computer Vision and Pattern Recognition
  (CVPR)}, 3973--3981.

\bibitem[{Ye and Bors(2021{\natexlab{a}})}]{LifelongTeacherStudent}
Ye, F.; and Bors, A. 2021{\natexlab{a}}.
\newblock Lifelong Teacher-Student Network Learning.
\newblock \emph{IEEE Transactions on Pattern Analysis and Machine
  Intelligence}.

\bibitem[{Ye and Bors(2020{\natexlab{a}})}]{LifelongVAEGAN}
Ye, F.; and Bors, A.~G. 2020{\natexlab{a}}.
\newblock Learning Latent Representations Across Multiple Data Domains Using
  Lifelong VAEGAN.
\newblock In \emph{Proc. of European Conference on Computer Vision (ECCV), vol.
  LNCS 12365}, 777--795.

\bibitem[{Ye and Bors(2020{\natexlab{b}})}]{LifelongInterpretable}
Ye, F.; and Bors, A.~G. 2020{\natexlab{b}}.
\newblock Lifelong learning of interpretable image representations.
\newblock In \emph{Proc. Int. Conf. on Image Processing Theory, Tools and
  Applications (IPTA)}, 1--6.

\bibitem[{Ye and Bors(2020{\natexlab{c}})}]{MixtureOfVAEs}
Ye, F.; and Bors, A.~G. 2020{\natexlab{c}}.
\newblock Mixtures of variational autoencoders.
\newblock In \emph{Proc. Int. Conf. on Image Processing Theory, Tools and
  Applications (IPTA)}, 1--6.

\bibitem[{Ye and Bors(2021{\natexlab{b}})}]{DeepMixtureVAE}
Ye, F.; and Bors, A.~G. 2021{\natexlab{b}}.
\newblock Deep Mixture Generative Autoencoders.
\newblock \emph{IEEE Transactions on Neural Networks and Learning Systems},
  1--15.

\bibitem[{Ye and Bors(2021{\natexlab{c}})}]{InfoVAEGAN_conference}
Ye, F.; and Bors, A.~G. 2021{\natexlab{c}}.
\newblock InfoVAEGAN: Learning Joint Interpretable Representations by
  Information Maximization and Maximum Likelihood.
\newblock In \emph{2021 IEEE International Conference on Image Processing
  (ICIP)}, 749--753.

\bibitem[{Ye and Bors(2021{\natexlab{d}})}]{JontLatentVAEs}
Ye, F.; and Bors, A.~G. 2021{\natexlab{d}}.
\newblock Learning joint latent representations based on information
  maximization.
\newblock \emph{Information Sciences}, 567: 216--236.

\bibitem[{Ye and Bors(2021{\natexlab{e}})}]{LifelongInfinite}
Ye, F.; and Bors, A.~G. 2021{\natexlab{e}}.
\newblock Lifelong Infinite Mixture Model Based on Knowledge-Driven Dirichlet
  Process.
\newblock In \emph{Proceedings of the IEEE/CVF International Conference on
  Computer Vision (ICCV)}, 10695--10704.

\bibitem[{Ye and Bors(2021{\natexlab{f}})}]{LifelongMixuteOfVAEs}
Ye, F.; and Bors, A.~G. 2021{\natexlab{f}}.
\newblock Lifelong Mixture of Variational Autoencoders.
\newblock \emph{IEEE Transactions on Neural Networks and Learning Systems},
  1--14.

\bibitem[{Ye and Bors(2021{\natexlab{g}})}]{LifelongTwin}
Ye, F.; and Bors, A.~G. 2021{\natexlab{g}}.
\newblock Lifelong Twin Generative Adversarial Networks.
\newblock In \emph{Proc. IEEE Int. Conf. on Image Processing (ICIP)},
  1289--1293.

\bibitem[{Yu and Grauman(2017)}]{shore_Dataset}
Yu, A.; and Grauman, K. 2017.
\newblock Semantic Jitter: Dense Supervision for Visual Comparisons via
  Synthetic Images.
\newblock In \emph{Proc. IEEE Int. Conf. on Computer Vision (ICCV)},
  5571--5580.

\bibitem[{Zenke, Poole, and Ganguli(2017)}]{Continual_Learning}
Zenke, F.; Poole, B.; and Ganguli, S. 2017.
\newblock Continual learning through synaptic intelligence.
\newblock In \emph{Proc. of Int. Conf. on Machine Learning (ICML), vol. PLMR
  70}, 3987--3995.

\end{thebibliography}
	}
	
\end{document}